\documentclass[runningheads]{llncs}

\usepackage{eccv}

\usepackage{eccvabbrv}

\usepackage{graphicx}
\usepackage{booktabs}

\usepackage[accsupp]{axessibility}  % Improves PDF readability for those with disabilities.

\usepackage{hyperref}

\usepackage{orcidlink}

\makeatletter  
\def\@fnsymbol#1{\ensuremath{\ifcase#1\or  
  \textrm{*}\or              % Symbol for first \thanks  
  \ddagger\or       % Symbol for second \thanks  
  \dagger\or       % Symbol for third \thanks  
  \mathparagraph\or % Etc.  
  \|\or  
  **\or  
  \dagger\dagger\or  
  \ddagger\ddagger\else\@ctrerr\fi}}  
\makeatother 
\usepackage{xcolor}
\usepackage{enumitem}
\usepackage{colortbl}
\usepackage{tcolorbox}
\usepackage{multirow}
\usepackage{multicol}
\usepackage{tcolorbox}
\usepackage{diagbox}
\usepackage{makecell} 
\usepackage{wrapfig,lipsum,booktabs}

\newcommand{\egno}{\textit{e}.\textit{g}.} %there is no space
\newcommand{\ieno}{\textit{i}.\textit{e}.} %there is no space
\newcommand{\ours}{VisualCritic}

\begin{document}

\title{VisualCritic: Making LMMs Perceive Visual Quality Like Humans} 

% \titlerunning{VisualCritic}

\author{
Zhipeng Huang\textsuperscript{\rm 1,2}\thanks{Equal Contributions.}\thanks{This work was done when Zhipeng Huang was an intern at MSRA.} \quad 
Zhizheng Zhang\textsuperscript{\rm 2*}\thanks{Corresponding Author.} \quad
Yiting Lu\textsuperscript{\rm 1} \quad
Zheng-Jun Zha\textsuperscript{\rm 1}\quad 
Zhibo Chen\textsuperscript{\rm 1}\quad 
Baining Guo\textsuperscript{\rm 2}\\
}

\authorrunning{Zhipeng et al.}

\institute{\textsuperscript{\rm 1}University of Science and Technology of China \quad 
\textsuperscript{\rm 2}Microsoft Research Asia \\
\email{\{hzp1104, luyt31415\}@mail.ustc.edu.cn} \quad \email{\{zhazj, chenzhibo\}@ustc.edu.cn} \\
\email{zhizzhangms@gmail.com} \quad
\email{bainguo@microsoft.com}
}
% Springer Heidelberg, Tiergartenstr.~17, 69121 Heidelberg, Germany

\maketitle

\begin{abstract}
  At present, large multimodal models (LMMs) have exhibited impressive generalization capabilities in understanding and generating visual signals. However, they currently still lack sufficient capability to perceive low-level visual quality akin to human perception. Can LMMs achieve this and show the same degree of generalization in this regard? If so, not only could the versatility of LMMs be further enhanced, but also the challenge of poor cross-dataset performance in the field of visual quality assessment could be addressed. In this paper, we explore this question and provide the answer "Yes!". As the result of this initial exploration, we present VisualCritic, the first LMM for broad-spectrum image subjective quality assessment. VisualCritic can be used across diverse data right out of box, without any requirements of dataset-specific adaptation operations like conventional specialist models. As an instruction-following LMM, VisualCritic enables new capabilities of (1) quantitatively measuring the perceptual quality of given images in terms of their Mean Opinion Score (MOS), noisiness, colorfulness, sharpness, and other numerical indicators, (2) qualitatively evaluating visual quality and providing explainable descriptions, (3) discerning whether a given image is AI-generated or photographic. Extensive experiments demonstrate the efficacy of VisualCritic by comparing it with other open-source LMMs and conventional specialist models over both AI-generated and photographic images.
\end{abstract}

%%%%%%%%% INTRODUCTION

\section{Introduction}

Research on Large Multimedia Models (LMMs) is at an exciting but still preliminary juncture. Current LMMs have shown impressive generalization capabilities in understanding high-level semantics \cite{alayrac2022flamingo,li2023blip,zhu2023minigpt,gpt4vision} and generating visual signals \cite{yu2023scaling,yu2023language,koh2023generating} in an instruction-following manner. Recent benchmark work \cite{wu2023q} and our experiments have both found that current LMMs struggle to perceive the low-level quality of visual signals in line with human perception. To remedy this shortcoming, we aim to comprehensively enhance the capability of LMMs in this aspect.
% towards achieving artificial general intelligence.

\begin{figure*}[!t]
	\centering
	\includegraphics[width=\textwidth]{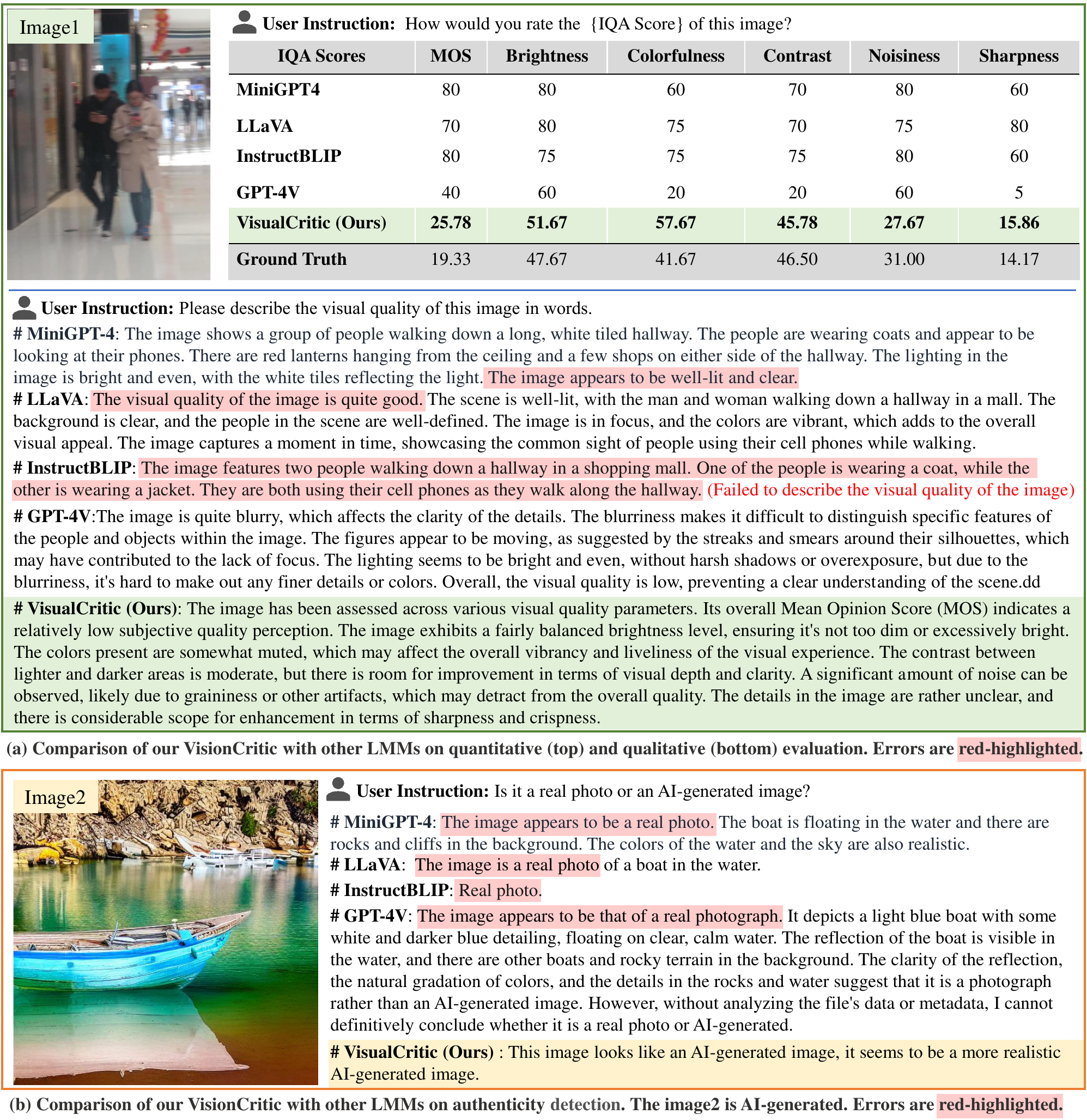}
        \vspace{-4mm}
    	\caption{Illustration of the comparison between existing LMMs and our proposed \ours~for visual subjective quality assessment from the perspectives of quantitative measurement, qualitative evaluation and authenticity detection. 
        % For each perspective, different models are tested with the same prompt to ensure the fairness. 
        The results show \ours~is the best one of its kind to perform consistently well over different relevant tasks.}
    % \vspace{-4mm}
	\label{fig:abstract_results}
\end{figure*}

Visual quality assessment aims to qualify the perceptual quality of visual signals in accordance with human perception. In this field, conventional specialist methods \cite{ghildyal2022stlpips,thong2022contentdiverse,roy2023test,saha2023re,zhang2023blind,zhao2023quality} have already achieved high statistical correlation with human subjective ratings for in-domain (within-dataset) settings. Although a given dataset can be well fitted, qualifying visual perceptual quality in a general scope remains highly challenging due to the diversity of visual signals and the complexity of human perception. Amidst the vast volume and diversity of visual signals we experience daily, the broad applicability of this technology determines its genuine practicality. Besides, we believe in its potential immense value in interplaying with AIGC models for better visual alignment, by serving as a generic reward model. Just as an excellent cook typically necessitates a sensitive palate. Thus, constructing a generally applicable visual quality assessment model is indisputably important. Beyond numerical results, we also expect a generalist model for visual quality assessment to enable more functionalities, \egno, explanatory descriptions or authenticity detection.
% The general-scope visual quality assessment is extremely important since it can capture human preference for visual signals and provide supervisory signals to achieve visual alignment for those  generation models. 
% Just as an excellent art creator first needs to have a high level of aesthetic judgment. 

Nevertheless, both current LMMs and conventional specialist models are far from satisfactory in this aspect. Regarding the \textbf{\textit{generalizability}}: The evaluation results in Figure \ref{fig:abstract_results} show that all LMMs, apart from ours, perform poorly in quantitative assessment, revealing clear gaps with humans ratings. Their basic quantitative assessment capabilities are lacking, let alone generalization. Conventional specialist models perform well for within-dataset visual quality assessment, but still struggle to cross-dataset evaluation due to unaligned human ratings over different datasets. They commonly rely on dataset-specific adaptation technologies \cite{li2021unified,zhu2020metaiqa,liu2022source}, requiring access to the target domain thus limiting their practicality.
Regarding the \textbf{\textit{versatility}}: As illustrated in Figure \ref{fig:abstract_results}, all open-source LMMs give the qualitative judgements obviously inconsistent with human perception. Their responses are more about semantic captions instead of the visual quality. GPT-4V \cite{gpt4vision} is able to give a basically reasonable response, but also tends to output semantic captions for scene understanding. All LMMs, apart from ours, fail to distinguish an AI-generated image. Conventional models do not supported these functionalities. 

% Nevertheless, the performance of current LMMs in this aspect is far from satisfactory. As our evaluation results illustrated in Figure \ref{fig:abstract_results}, apart from ours, other LMMs all struggle to precisely qualify the perceptual quality for the given image, and also fail to discern whether the given image is photographic or AI-generated. As for the qualitative evaluation responses, among the LMMs for comparison, only GPT-4V \cite{gpt4vision} is able to give reasonable responses while others cannot.

In this work, we make the first endeavour to build an LMM for broad-spectrum image subjective quality assessment, taking into account both generalizability and versatility. To this end, we find the model architecture is not the key, but an appropriate data construction and an effective training strategy are. We thus follow the common practices in building other LLMs \cite{li2023blip,zhu2023minigpt,peng2023kosmos,chen2023minigpt,instructblip,liu2023visual} to configure the model with a pre-trained vision encoder, a pre-trained LLM and a learnable adapter. Without a doubt, collecting a large-scale dataset with human ratings from scratch is extremely expensive. A straightforward solution is to combine publicly accessible datasets into a big one for joint training. However, in fact, this fails to deliver favorable results as we ever imagined, due to the rating inconsistency among different datasets. Detailed reasons are analyzed later. Through data analysis, we realize that the relativity of human ratings is much more transferable across different datasets than their absolute values. To utilize this characteristic, we propose a multi-stage curriculum learning strategy, in which we first perform the relativity learning on large-scale data from diverse sources and then adopt a few samples to unleash the capability of precisely estimating absolute scores. This effectively facilitates the learning of enabling an LMM to perceive low-level visual quality in a broad range akin to human perception. 

As a result, we propose \ours, the first LMM of its kind to support broad-spectrum image subjective quality assessment. Compared to conventional specialist models in the field of visual quality assessment, \ours~exhibits the state-of-the-art cross-dataset generalization ability over both photographic and AI-generated images. Besides, in addition to chat capabilities, it enables more domain-relevant functionalities for providing qualitative descriptions and performing authenticity detection.

\section{Related Work}

% [To be considered] Is it needed to include those works related to 3D or other modalities? 
\subsection{Large Multimodal Models}
Recently, Large Multimodal Models (LMMs) have begun to showcase their preliminary prowess, riding on the coattails of the success of Large Language Models (LLMs). Initial research efforts \cite{alayrac2022flamingo,li2023blip,zhu2023minigpt} in this field have managed to accomplish preliminary multimodal universality by merely training a minimal number of parameters serving for adaptation and projection, effectively uniting frozen visual foundation models with pre-trained language foundation models. Subsequent research works further enhance their capabilities from different perspectives. Kosmos-2 \cite{peng2023kosmos}, MiniGPT-v2 \cite{chen2023minigpt} and Ferret \cite{you2023ferret} enhance the spatial perception capabilities of multimodal large models, equipping them to be more applicable for a wide array of tasks with object bounding boxes involved. Kosmos-2.5 \cite{lv2023kosmos} develops the literate capability. Kosmos-G \cite{pan2023kosmos}, MiniGPT-5 \cite{zheng2023minigpt} and CM3Leon \cite{yu2023scaling} try to unify both general-purpose semantic understanding and visual generation with a single model. Besides, there are a series of works building multimodal agents \cite{AutoGPT,AgentGPT,yang2023mm,shen2023hugginggpt,zhang2023responsible} to systematically complete their functionality and enhance their practicality in different application scenarios. In this work, we make the first endeavour towards a leap in the low-level perception capability of LMMs for visual quality. 

% We believe this could enable a comprehensive perception including both low-level quality and high-level semantics of visual signals, as well as its mutual benefits for the advancement in visual generation capabilities.

\subsection{Visual Quality Assessment}

Visual quality assessment aims to quantify the perceptual quality of visual signals in accordance with human perception, beyond classic metrics like PSNR or SSIM \cite{wang2004image}. Full-reference Image Quality Assessment (IQA) \cite{ghildyal2022stlpips,thong2022contentdiverse} requires estimating the quality distance of the distorted image between the reference image to then estimate the quality score of the distorted one. In contrast, a more challenging and practical task setting is the no-reference image quality assessment \cite{roy2023test,saha2023re,zhang2023blind,zhao2023quality}, also known as blind image quality assessment, wherein no reference images are provided. In this field, it is relatively easy to fit a given dataset while the real challenge lies in being generally applicable for samples in the wild. For this, cross-dataset adaptation technologies \cite{li2021unified,zhu2020metaiqa,liu2022source} are developed. But they require the annotations of target datasets for model fine-tuning, which limits their practicality and imposes strong application and research demands for developing an IQA generalist model that could be used out of the box on diverse data. 
% with strong generalization ability. 

With the advent of large-scale models, GPT-4V \cite{gpt4vision} has been explored to evaluate text-to-image results from the perspective of design quality in \cite{lin2023designbench} and shows very preliminary promise in predicting human preference for visual aesthetics and text-image alignment \cite{yang2023dawn}. For IQA, as we evaluated, it can provide basically reasonable textual descriptions, but cannot accurately predict quantitative results or distinguish between AI-generated images and real images. Besides, a series of open-source LMMs are evaluated right out of the box in \cite{wu2023q} but they all perform poorly on quantitative visual quality assessment.
% compared to conventional IQA methods. 
We are the first to construct an LMM for IQA with unprecedented generalizability and versatility. In addition to quantitative assessment as prior specialist models, our model can further provide linguistic descriptions and accomplish authenticity detection.

%%%%%%%%% METHOD

\begin{figure*}[!t]
	\centering
    \vspace{-4mm}
	\includegraphics[width=0.96\textwidth]{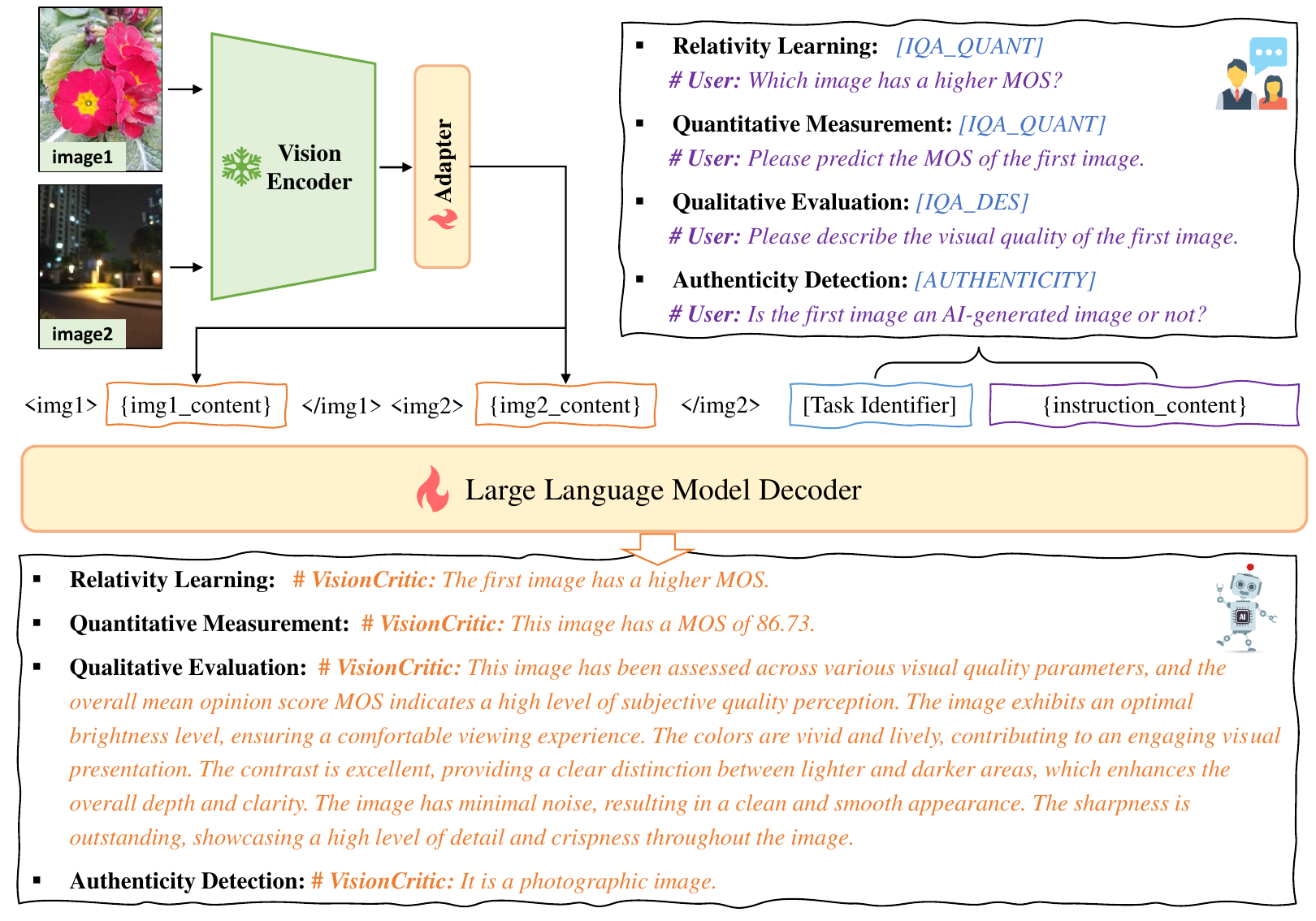}
	\caption{The framework of our proposed \ours, which comprises a frozen vision encoder, a learned cross-modality adapter and a LoRA-tuned LLM decoder. \ours~is the first of its kinds to support diverse visual quality assessment tasks, including relative quality comparison, quantitative measurement, qualitative evaluation and authenticity detection.}
	\label{fig:framework}
 % \vspace{-8mm}
\end{figure*}

\section{\ours}

% We introduce an LMM for broad-spectrum visual quality assessment, named \ours. We introduce its architecture in Sec.\ref{sec:architecture}, then elaborate the data construction method and the training strategy in Sec.\ref{sec:data_construction} and Sec.\ref{sec:training_strategy}, respectively.

\subsection{Model Architecture}
\label{sec:architecture}

As indicated in recent reports \cite{ghildyal2022stlpips,thong2022contentdiverse}, current LMMs have preliminarily shown the potential to capture the perceptual quality of images, but they still cannot make precise judgments and support comprehensive assessment in the general scope. For our \ours, we inherit as much world knowledge as possible and the conversational ability from the pre-trained weights and infuse new knowledge to endow it with our target capability. 
As illustrated in Figure \ref{fig:framework}, \ours~consists of a vision encoder, a LLM decoder and a cross-modality adapter as common designs in \cite{zhu2023minigpt,li2023blip,peng2023kosmos,lv2023kosmos,chen2023minigpt}. In specific, we adopt a pre-trained EVA model \cite{fang2023eva} as the vision encoder to encode vision inputs into a series of tokens. These tokens are transformed to the input embedding space of a LLM decoder by an adapter with the same architecture as that in \cite{li2023blip}. We adopt an open-sourced Vicuna-13B \cite{chiang2023vicuna} as the LLM decoder. We empirically find the model architecture is not the key to \ours, instead, the data construction and the training strategy are. We delineate them in the following.

\subsection{Data Construction}
\label{sec:data_construction}

Given an image $x$, \ours~could assess its perceptual quality by not only estimating a series of quantitative metrics $y_{score}$ like conventional specialist models but also providing natural language descriptions $y_{desc}$. This advances explainable visual quality assessment. Besides, it can also distinguish whether a given image $x$ is AI-generated or not, outputting $y_{aigc}$. We introduce their corresponding data construction respectively as follows.

% \vspace{-3mm}
\noindent\textbf{Quantitative measurement.} \ours~quantitatively assesses the perceptual quality of given images by estimating the most commonly used metric for subjective quality, \ieno, Mean Opinion Score (MOS) \cite{streijl2016mean}, and a series of quality attribute indicators including the brightness, colorfulness, contrast, noisiness, and sharpness. To enable this, we gather data with human subjective ratings from different public sources (details are listed in Table \ref{tab:dataset_and_training}) and normalize the numerical annotations to the range of $[0, 100]$. 

Nevertheless, training a visual quality assessment generalist with mixed datas-ets is much more challenging than we ever imagined. We experimentally find that straightforwardly combining samples from different datasets in joint training even performs worse than training on a single dataset when performing cross-dataset evaluation. This is because subjective annotations of different datasets are collected in different experiment environments and from different annotators \cite{li2021unified}. Due to the inherent differences in visual quality among different datasets, it is difficult for different annotators to score a certain dataset with a unified standard over all datasets. \textbf{Such inconformity starves generalization.}

An intuitive way to address this inconformity issue is to rebuild a large-scale dataset over diverse images and have annotators score with a unified standard, which is, however, quite costly and hard to accomplish. Considering that the judgement of relative quality is obviously easier to generalize, 
% Considering that the judgement of relative quality is free of affected across datasets, 
we propose a curriculum learning scheme.
% for quantitative measurement. 
Specifically, we first teach \ours~to discern the relativity of visual quality using large-scale data from diverse sources, then use a few samples without the label inconformity issue to enable it to further output absolute scores. We take MOS as an example to introduce the data construction process in detail, with other metrics employing analogous methods. For the relativity learning, we sample two different images $\{x_1, x_2\}$ from the same dataset each time, and set the corresponding label $y_{mos}\!=\![1,0]$ if the MOS of $x_1$ is higher than that of $x_2$, otherwise $y_{mos}\!=\![0,1]$. Then, we prompt GPT-4 \cite{gpt4} to generate the dialogue data upon the relative quality labels following the conversation template design of Vicuna \cite{chiang2023vicuna}. For the absolute score estimation, we take the normalized MOS as its corresponding labels, and structure the training data in a dialogue form, following a shared format used for relativity learning. This unified format will be detailed in the following. 

% \vspace{-3mm}
\noindent\textbf{Qualitative evaluation.} \ours~can further provide qualitative evaluation regarding the perceptual quality of given images in a more comprehensive view. The outputs incorporate but are not limited to various indicators in quantitative measurement, beyond their numerical values, enabling interpretability. To achieve this, given each sample, we prompt GPT-4 \cite{gpt4} to generate its corresponding dialogue-form training data in the same format with that for quantitative measurement on top of the ground truths of aforementioned quantitative indicators. The knowledge of GPT-4 is distilled in the annotations of our training data in this way. As a result, \ours~integrates the knowledge pertinent to visual quality assessment from its own pre-trained weights, GPT-4 and multiple public datasets, making it be a model ``standing on the shoulders of giants''. % The adopted prompts for annotation generation are not the focus of this paper and are detailed in the supplementary.

% \vspace{-3mm}
\noindent\textbf{Authenticity detection.} Aforementioned functionalities of quantitative measurement and qualitative evaluation can be applied for both AI-generated and photographic images. Towards further versatility, \ours~is also endowed with the authenticity detection capability, \ieno, discerning whether a given image is AI-generated or photographic. For a given image $x$, its label regarding the authenticity $y_{auth}\!=\!1$ if $x$ is a photographic image, otherwise $y_{auth}\!=\!0$. $(x, y_{auth})$ is prompted to be a dialogue form, analogous to those for quantitative measurement and qualitative evaluation. Details are in the supplementary.

% \vspace{-3mm}
\noindent\textbf{Unified format and data summary.} The training data for above three functionalities adopt a unified format following the conversation template design of Vicuna \cite{chiang2023vicuna} as below:
\begin{tcolorbox}
\textbf{Human:} $<img1>\{img1\_content\}</img1>(<\!img2\!>\{img2\_content\}<\!/img2\!>)[Task\;\\Identifier]\{instruction\_content\}$\\
\textbf{Assistant:} $\{answer\_content\}$
\label{box:unified_format}
\end{tcolorbox}
\noindent~In this format, $<img>$ and $</img>$ are special tokens marking the start and end of the token sequence of an image, respectively. $\{img\_content\}$, $\{instruction\_content\}$ and $\{answer\_content\}$ are three placeholders of the tokens of input image, user instruction and \ours's response, respectively. The contents in the brackets are optional, only existing in the training data of the relativity learning, not in others. Inspired by \cite{chen2023minigpt}, we treat the learning for aforementioned three functionalities as a multi-task problem and introduce a task identifier $[Task\;Identifier]$ to reduce the ambiguity across different tasks. It is instantiated as $<IQA\_QUANT>$, $<IQA\_DES>$ and $<AUTHENTICITY>$ for quantitative measurement, qualitative evaluation and authenticity detection, respectively. We summarize all adopted datasets for different purposes in Table \ref{tab:dataset_and_training}, and detail their corresponding training strategies subsequently.

\subsection{Training Strategy}
\label{sec:training_strategy}

To ensure both high generalization performance and versatility, we introduce a multi-stage training strategy for \ours, with each stage serving its own purpose as shown in Table \ref{tab:dataset_and_training}. Let us detail them below.

\begin{table*}[t]
  \centering
  \caption{Details of the datasets used for different training stages. \textit{Relativity Learning} refers to the learning of ranking two images according to their subjective quality. \textit{Quant.} denotes the training for enabling \ours~to quantitatively score images. \textit{Quali.} represents the training for enabling qualitative evaluation.}
  \resizebox{\linewidth}{!}{
    \begin{tabular}{l|l|ccc}
    \toprule
    \multicolumn{1}{c|}{Purposes} & \multicolumn{1}{c|}{Datasets} & Stage-1 & Stage-2 & Stage-3 \\
    \midrule
    % Relativity Learning & \makecell[lt]{KonIQ-10k \cite{hosu2020koniq}, SPAQ \cite{fang2020perceptual}, FLIVE \cite{ying2020patches}, LIVEC \cite{ghadiyaram2015massive},\\ CGIQA-6K \cite{zhang2023subjective}, AGIQA-3k \cite{li2023agiqa}, KADID-10k \cite{lin2019kadid}} & \checkmark & \checkmark & \checkmark \\
    Relativity Learning & \makecell{KonIQ-10k \cite{hosu2020koniq}, SPAQ \cite{fang2020perceptual}, FLIVE \cite{ying2020patches}, LIVEC \cite{ghadiyaram2015massive},\\ CGIQA-6K \cite{zhang2023subjective}, AGIQA-3k \cite{li2023agiqa}, KADID-10k \cite{lin2019kadid}} & \checkmark & \checkmark & \checkmark \\
    % \cline{2-5}
    % \hdashline[0.5pt/5pt]
    \hline
    Quant. \& Quali. & KonIQ-10k \cite{hosu2020koniq} &       & \checkmark & \checkmark \\
    % \cline{2-5}
    % \hdashline[0.5pt/5pt]
    \hline
    Authenticity Detection & KonIQ-10k \cite{hosu2020koniq}, AIGCIQA-2023 \cite{wang2023aigciqa2023}  &       & \checkmark & \checkmark \\
    % \cline{2-5}
    % \hdashline[0.5pt/5pt]
    \hline
    Instruction Tuning & LLaVA-Instruction \cite{liu2023visual}, CCSBU-Aligned \cite{zhu2023minigpt} &       &       & \checkmark \\
    \bottomrule
    \end{tabular}%
    }
  \label{tab:dataset_and_training}%
\end{table*}%

% \vspace{-3mm}
\noindent\textbf{Stage-1: relativity learning.} We first subject \ours~to the relativity learning with large-scale data from different sources for enabling it to rank the relative quality of multiple image diversely. This can effectively address the label inconformity issue on visual subjective quality across different datasets and is consistent with the intuition that it is easier to rank image of different quality than to rate them. In this stage, \ours~acquires preliminary perception for visual quality aligned with human preference.

% \vspace{-3mm}
\noindent\textbf{Stage-2: multifunctional training.} Afterwards, we perform multifunctional training to unleash the capabilities of quantitative measurement, qualitative evaluation and authenticity detection as introduced before. We integrate the data for different functionalities as described in Sec. \ref{sec:data_construction} into a joint training framework, where we use $<Task\;Identifier>$ to differentiate tasks and each batch only contains one. In this stage, the knowledge of \ours~acquired in the stage-1 are further enhanced to accomplish specific tasks in an instruction following manner.

% \vspace{-3mm}
\noindent\textbf{Stage-3: instruction tuning.} Finally, we manually select some high-quality samples from the training data of the first two stages and combine them with general-purpose dialogue data to perform the instruction tuning for \ours, so as to improving the quality and robustness of its responses to user instructions.

The data sources for the training of different stages are summarized in Table \ref{tab:dataset_and_training}. They are all structured to be the aforementioned unified dialogue format in the way described in Sec. \ref{sec:data_construction}. This modelling allows us to train \ours~with a plain but universal auto-regressive optimization objective, which can be denoted as below:
\begin{equation} \label{eq:1}
    maximize\; \sum_{n=2}^{N_y} \mathbf{E}_{\hat{P}}[\log P(\boldsymbol{y}_n | \boldsymbol{x}_{1:N_x}, \boldsymbol{y}_{1:n-1})],
\end{equation}
where $\mathbf{E}_{\hat{P}}[\cdot]$ is the expectation operator with respect to the distribution $\hat{P}$. Here, $\hat{P}$ is the expected distribution (\ieno, ground-truth distribution) of $P$. $\mathbf{E}_{\hat{P}}[\cdot]$ is commonly implemented by a cross-entropy function between $P$ and $\hat{P}$. $\boldsymbol{x}_{1:N_x}$ and $\boldsymbol{y}_{1:n-1}$ are the image tokens and preceding language tokens, respectively.

%%%%%%%%% EXPERIMENTS

\section{Experiments}

\subsection{Implementation Details}
\label{sec:exp_details}
Throughout the training process, the visual backbone of~\ours~remains frozen. The linear layer in the adapter and the language model are initialized with the weights of \cite{zhu2023minigpt} and are trained efficiently with LoRA~\cite{hu2022lora,peft} techniques.
% Our focus is on developing the linear projection layer and refining the language model efficiently through LoRA~\cite{hu2022lora,peft} techniques. 
We employ a rank setting of r = 16. The model is trained using a $224\times224$ image resolution at all stages. We employ the AdamW~\cite{loshchilov2017decoupled} optimizer paired with a cosine learning rate scheduler~\cite{loshchilov2016sgdr} for model training. In the first stage, we conduct training over 50,000 steps using 4$\times$A100 GPUs, with a global batch size of 96 and a base learning rate of 2e-4. This stage is completed in approximately 12 hours. In the second stage, the model is trained for 10,000 steps on 4$\times$A100 GPUs, maintaining a global batch size of 96 and a base learning rate of 5e-5, taking around 2 hours. Finally, the model undergoes an additional 5,000 training steps on 4$\times$A100 GPUs, with a global batch size of 64, completed in about 1 hour. The maximum learning rate remains constant at 1e-5 during this final stage.

We normalize the ranges of human ratings with linear scaling to [0-100]. Note that the rating inconsistencies in specific values persist as analyzed in Sec.~\ref{sec:data_construction}. This is the driving force behind our proposal of relativity learning.

Unless otherwise stated, we employ Spearman Rank Correlation Coefficient (SRCC) and Pearson's Linear Correlation Coefficient (PLCC) as the evaluation metrics following the common practices in this field \cite{wu2023q,roy2023test,saha2023re,zhang2023blind}. More implementation details are provided in supplementary.

\subsection{Quantitative Results}
\label{sec:exp_quantitative}

% We analyze the quantitative performance of \ours, mainly focusing on the most widely used MOS, including other indicators as well.

\begin{table*}[t]
  \centering
  \caption{Comparison results (MOS) of our proposed \ours~with other LMMs on zero-shot visual subjective quality assessment. Metrics are shown as SPCC/PLCC. Ins-blip$^1$ is the abbreviation for InstructBLIP (Vicuna), and Ins-blip$^2$ stands for InstructBLIP (T5), the baseline of \ours~ is MiniGPT4~\cite{zhu2023minigpt}}.
  \resizebox{\linewidth}{!}{
    \begin{tabular}{l|ccccccc}
    \toprule
    Models   & KonIQ-10k & SPAQ & FLIVE & LIVEC & CGIQA-6K & KADID-10k & AGIQA-3k  \\
    \midrule
    CLIP~\cite{radford2021learning} & 0.497/0.532 & 0.351/0.358 & 0.214/0.236 & 0.381/0.360 & 0.356/0.344 & 0.395/0.408 & 0.392/0.433 \\
    \hline
    Llava-v1~\cite{liu2023visual} & 0.494/0.500 & 0.426/0.459 & 0.258/0.290 & 0.363/0.391 & 0.054/0.088 & 0.374/0.401 & 0.601/0.669 \\
    Kosmos-2~\cite{peng2023kosmos} & 0.280/0.305 & 0.653/0.649 & 0.204/0.204 & 0.377/0.380 & 0.207/0.215 & 0.358/0.368 & 0.516/0.506 \\
    InternLM~\cite{zhang2023internlm} & 0.587/0.635 & 0.738/0.753 & 0.359/0.413 & 0.573/0.649 & 0.176/0.172 & 0.533/0.567 & 0.711/\textbf{0.772} \\
    Mplugowl~\cite{ye2023mplug} & 0.476/0.482 & 0.645/0.650 & 0.236/0.273 & 0.367/0.403 & 0.094/0.103 & 0.473/0.495 & 0.666/0.698 \\
    Qwen-vl~\cite{bai2023qwen} & 0.501/0.570 & 0.673/0.664 & 0.306/0.346 & 0.499/0.509 & 0.212/0.211 & 0.505/0.500 & 0.595/0.676 \\
    Shikra~\cite{chen2023shikra} & 0.352/0.340 & 0.327/0.334 & 0.232/0.237 & 0.244/0.278 & 0.155/0.139 & 0.325/0.340 & 0.622/0.645 \\
    % Llama-ada2~\cite{gao2023llama} & 0.386/0.389 & 0.472/0.514 & 0.276/0.327 & 0.274/0.339 & 0.224/0.234 & 0.433/0.441 & 0.551/0.619 \\
    Llava-v1.5~\cite{liu2023visual} & 0.471/0.487 & 0.547/0.575 & 0.304/0.337 & 0.372/0.438 & 0.306/0.313 & 0.417/0.432 & 0.625/0.748 \\
    % Visualglm & 0.274/0.243 & 0.501/0.507 & 0.159/0.173 & 0.052/0.047 & 0.121/0.116 & 0.129/0.133 & 0.300/0.308 \\
    Idefics~\cite{laurenccon2023obelics} & 0.405/0.434 & 0.468/0.488 & 0.234/0.243 & 0.411/0.438 & 0.179/0.201 & 0.368/0.383 & 0.551/0.614 \\
    Ins-blip$^1$~\cite{instructblip} & 0.383/0.458 & 0.688/0.689 & 0.189/0.268 & 0.206/0.330 & 0.198/0.230 & 0.343/0.387 & 0.611/0.665 \\
    Ins-blip$^2$~\cite{instructblip} & 0.312/0.307 & 0.602/0.632 & 0.219/0.220 & 0.000/0.039 & 0.066/0.110 & 0.281/0.240 & 0.235/0.255 \\
    Otter-v1~\cite{li2023otter} & 0.421/0.424 & 0.478/0.484 & 0.140/0.139 & -0.020/0.030 & 0.336/0.363 & 0.568/0.587 & 0.429/0.460 \\
    \midrule
    baseline~\cite{zhu2023minigpt} & 0.276/0.289 & 0.250/0.275 & 0.180/0.199 & 0.315/0.314 & 0.224/0.199 & 0.248/0.246 & 0.572/0.584 \\
    \rowcolor[gray]{0.9}\textbf{\ours} & \textbf{0.856/0.867} & \textbf{0.844/0.852} & \textbf{0.553/0.633} & \textbf{0.738/0.778} & \textbf{0.442/0.473} & \textbf{0.681/0.701} & \textbf{0.734}/0.712 \\
    \bottomrule
    \end{tabular}%
    }
  \label{tab:quant_lmms}%
\end{table*}%

\noindent\textbf{Comparison to LMMs.} Our goal is to build an LMM that can perceive visual quality like humans. We thus first compare the proposed \ours~to other open-source LMMs in this aspect. The alignment between model outputs and human preference on the most widely used metric MOS is measured by two correlation coefficients SPCC and PLCC in Table \ref{tab:quant_lmms}. 
% These results are tested in the zero-shot setting in the sense that there are not any instance examples appearing in their prompts. 
We can observe that across different datasets including both photographic and AI-generated ones, \ours~performs the best and outperforms other LLMs by a large margin. This demonstrates that our \ours~statistically achieves the highest consistency with human preference of visual perceptual quality among existing LMMs, whether for photographic or AI-generated images. It showcases a high potential for improved visual alignment. We selected the MiniGPT4~\cite{zhu2023minigpt} as our baseline due to its straightforward implementation, allowing us to eliminate the confounding effects of more elaborate designs on our experimental results. A comparison of it with other LLMs is presented in Table~\ref{tab:quant_lmms}. Moreover, it is impractical to require LLMs in Table~\ref{tab:quant_lmms} to be compared using the same training data since their pre-training and training are performed on large-scale general-purpose datasets. These datasets may already encompass the data for IQA. Additionally, the data construction is a key factor in determining the varying capabilities of different large models, which also stands as one of our main contributions. 

We admit that there is still a performance gap between \ours~and SOTA IQA specialist models in terms of the within-dataset quantitative performance. Detailed comparison results can be found in our supplementary. The significance of developing \ours, such an LMM for this lies in its generalizability and practicality. \ours~attains the performance over different datasets listed in Table \ref{tab:quant_lmms} with a single model right out of the box, while a model is trained for a specific dataset for conventional IQA methods. The latter kind fits a given dataset well but suffers from poor cross-dataset generalization, severely limiting the practicality. To the best of our knowledge, \ours~is the first one of its kind to handle such diverse datasets including both photographic and AI-generated samples simultaneously. It can be constructed upon a commonly used LMM architecture, trained with a plain auto-regressive loss and can be used effortlessly right out of the box without additional adaptation. Besides, \ours~provides rich functionalities as introduced before and has the chat ability.
% the ability to chat with users.


% \begin{wraptable}{r}{0.62\textwidth}
\begin{table}[t]
  \centering
  % \vspace{-14mm}
  \caption{Comparison results (MOS) of our \ours~with conventional specialist models on the generalizable visual subjective quality assessment. The ``\textit{S}'' and ``\textit{T}'' denotes the source and target datasets, respectively. For \textit{\ours-S}, we only perform the training of Stage-2 and Stage-3, corresponding to a single-source generalization. For \textit{\ours-M}, we perform the training of all three stages with the target dataset discarded from the training data in all stages, corresponding to a multi-source generalization. Metrics are shown as SPCC/PLCC.}
  \resizebox{0.62\textwidth}{!}{
    \begin{tabular}{l|cccc}
    \toprule
    \diagbox{\small{Models}}{\small{S$\rightarrow$T}}  & \makecell{KonIQ-10k$\rightarrow$\\KADID-10k} & \makecell{KADID-10k$\rightarrow$\\KonIQ-10k} & \makecell{LIVEC$\rightarrow$\\KADID-10k} & \makecell{KADID-10k$\rightarrow$\\LIVEC} \\
    \midrule
    DBCNN~\cite{zhang2018blind} & 0.455/0.464 & 0.413/0.421 & 0.419/0.419 & 0.266/0.290 \\
    HyperIQA~\cite{su2020blindly} & 0.511/0.516 & 0.545/0.556 & 0.378/0.439 & 0.490/0.487 \\
    RankIQA~\cite{liu2017rankiqa} & 0.487/0.426 & 0.603/0.551 & 0.416/0.390 & 0.491/0.495 \\
    MUSIQ~\cite{ke2021musiq} & 0.486/0.504 & 0.554/0.573 & 0.497/0.524 & 0.517/0.524 \\
    GraphIQA~\cite{10.1109/TMM.2022.3152942} & 0.521/0.536 & 0.427/0.430 & 0.365/0.393 & 0.388/0.407 \\
    CLIP-IQA~\cite{wang2022exploring} & 0.655/0.655 & -  & -  & -  \\
    VCRNet~\cite{pan2022vcrnet} & 0.504/0.517 & 0.566/0.585 & 0.340/0.416 & 0.520/0.530 \\
    UCDA~\cite{ganin2016domain}  & 0.430/0.437 & 0.638/0.636 & 0.383/0.432 & 0.499/0.484 \\
    RankDA~\cite{chen2021unsupervised} & 0.403/0.413 & 0.496/0.501 & 0.420/0.485 & 0.583/0.619 \\
    \midrule
    \rowcolor[gray]{0.9}\textbf{\ours-S} & 0.656/0.668 & 0.657/0.676 & 0.559/0.533 & 0.645/0.646 \\
    % \rowcolor[gray]{0.9}\textbf{\ours-S} & \textbf{0.626/0.641} & \textbf{0.653/0.664} & \textbf{0.538/0.553} & \textbf{0.652/0.658} \\
    % \midrule
    % \diagbox{\small{Models}}{\small{S$\rightarrow$T}}  & \makecell{-} & \makecell{MultiSource$\rightarrow$\\KonIQ-10k} & \makecell{MultiSource$\rightarrow$\\KADID-10k} & \makecell{MultiSource$\rightarrow$\\LIVEC} \\
    \midrule
    \rowcolor[gray]{0.9}\textbf{\ours-M} & \textbf{0.679/0.683} & \textbf{0.665/0.686} & \textbf{0.562/0.571} & \textbf{0.661/0.671} \\
    \bottomrule
    \end{tabular}%
    }
  \label{tab:quant_conventional}%
\end{table}%
% \end{wraptable}

% \vspace{-3mm}
\noindent\textbf{Comparison to conventional IQA models.} We further compare our \ours~with conventional IQA specialist models in terms of their cross-dataset generalization ability. Detailed results are in Table \ref{tab:quant_conventional}. For single-source generalization, \textit{\ours-S} is superior to conventional IQA models thanks to the integration of world knowledge from pre-trained weights (via parameter inheritance) and GPT-4 (via annotation refinement/generation). Regarding multi-source generalization, the proposed \textit{\ours-M} performs better on the target datasets, attaining 0.024/0.028, 0.062/0.050, 0.065/0.047, 0.078/0.052 improvements for the \textit{KonIQ-10k$\rightarrow$KADID-10k}, \textit{KADID-10k$\rightarrow$KonIQ-10k}, \textit{LIVEC$\rightarrow$KA\\DID-10k}, \textit{KADID-10k$\rightarrow$LIVEC} settings on SRCC and PLCC, respectively, relative to the second best conventional IQA models. Compared to \textit{\ours-S}, the larger improvements achieved by \textit{\ours-M} demonstrate the effectiveness of our adopted relativity learning on the joint utilization of different IQA datasets.

\begin{table}[t]
  \caption{Quantitative results on other indicators. For brightness, colorfulness, contrast, noisiness and sharpness, we report their corresponding SPCC/PLCC on the SPAQ \cite{fang2020perceptual} dataset that provides the ground truths for these indicators. For authenticity detection, we report the accuracy on the photographic dataset SPAQ \cite{fang2020perceptual} and the AI-generated dataset AIGCIQA-2023 \cite{wang2023aigciqa2023}.}
  \centering
    \resizebox{0.62\linewidth}{!}{
    \begin{tabular}{cccccc}
    \toprule
    Brightness & Colorfulness & Contrast & Noisiness & Sharpness & Authenticity \\
    % Brightness (SRCC/PLCC) & Colorfulness (SRCC/PLCC) & Contrast (SRCC/PLCC) & Noisiness (SRCC/PLCC) & Sharpness (SRCC/PLCC) & Authenticity (Acc \%) \\
    \midrule
    \rowcolor[gray]{0.9} 0.730/0.783 & 0.707/0.719 & 0.709/0.717 & 0.813/0.807 & 0.828/0.811 & 70.3\% \\
    \bottomrule
    \end{tabular}%
    }
  % \vspace{-3mm}
  \label{tab:results_other_indicators}%
\end{table}%

% \vspace{-3mm}
\noindent\textbf{Quantitative results on other indicators.} Besides the most commonly used indicator, \ieno, MOS, we also report the quantitative results of our \ours~on other indicators in Table \ref{tab:results_other_indicators} to show its achieved versatility. Few publications report their performance on these indicators.

\begin{table*}[htbp]
  \centering
  \caption{Comparison results (MOS) of different training strategies for building \ours. \textit{MultiFunc$_{joint}$} refers to performing multifunctional training jointly over multiple datasets. \textit{MultiFunc$_{single}$} refers to performing multifunctional training on a single dataset. (Here, we use KonIQ-10k.) \textit{Relat.} is short for ``Relativity Learning'' for learning to rank multiple images according to their subjective quality. For all models in this table, the stage-3 training (instruction tuning) has been conducted, which is omitted for brevity. Metrics are shown as SPCC/PLCC}
  \resizebox{\linewidth}{!}{
    \begin{tabular}{l|ccccccc}
    \toprule
    \diagbox{\small{Methods}}{\small{\quad Datasets}}  & KonIQ-10k & SPAQ & FLIVE & LIVEC & CGIQA-6K & KADID-10k & AGIQA-3k \\
    \midrule
    MultiFunc$_{joint}$ & 0.621/0.664 & 0.701/0.703 & 0.247/0.197 & 0.476/0.485 & 0.130/0.141 & 0.337/0.263 & 0.477/0.484 \\
    MultiFunc$_{single}$ & 0.787/0.803 & 0.771/0.782 & 0.373/0.423 & 0.652/0.658 & 0.385/0.377 & 0.626/0.641 & 0.663/0.644 \\
    Relat. + MultiFunc$_{joint}$ & 0.643/0.682 & 0.726/0.716 & 0.251/0.203 & 0.479/0.513 & 0.171/0.183 & 0.384/0.333 & 0.525/0.537 \\
    \midrule
    \rowcolor[gray]{0.9} Relat. + MultiFunc$_{single}$ & \textbf{0.856/0.867} & \textbf{0.844/0.852} & \textbf{0.553/0.633} & \textbf{0.738/0.778} &  \textbf{0.442/0.473} & \textbf{0.681/0.701} & \textbf{0.734/0.712} \\
    \bottomrule
    \end{tabular}%
  }
  \label{tab:ablation_training}%
\end{table*}%

\begin{table*}[t]
  \centering
    \caption{Comparison results (MOS) of different dataset choices for the training of quantitative measurement in Stage-2 (multifunctional training). The datases used for the training of Stage-1 and Stage-3 remain the same as reported in Table \ref{tab:dataset_and_training}. Metrics are shown as SPCC/PLCC}
  \resizebox{\linewidth}{!}{
    \begin{tabular}{l|ccccccc}
    \toprule
    \diagbox{\small{Source}}{\small{Target}}   & KonIQ-10k & SPAQ & FLIVE & LIVEC & CGIQA-6K & KADID-10k & AGIQA-3k  \\
    \midrule
    KonIQ-10k~\cite{hosu2020koniq} & \textbf{0.856/0.867} & 0.834/0.849 & 0.551/0.634 & 0.735/0.775 & 0.458/0.473 & 0.706/0.709 & 0.739/0.736 \\
    SPAQ~\cite{fang2020perceptual} & 0.731/0.762 & \textbf{0.865/0.864} & 0.487/0.511 & 0.718/0.716 & 0.387/0.406 & 0.502/0.536 & 0.721/0.700 \\
    FLIVE~\cite{ying2020patches} & 0.707/0.717 & 0.857/0.853 & \textbf{0.633/0.667} & 0.754/0.809 & 0.252/0.295 & 0.521/0.558 & 0.639/0.625 \\
    LIVEC~\cite{ghadiyaram2015massive} & 0.760/0.747 & 0.807/0.822 & 0.582/0.580 & \textbf{0.857/0.864} & 0.399/0.398 & 0.578/0.591 & 0.645/0.684 \\
    CGIQA-6K~\cite{zhang2023subjective} & 0.525/0.535 & 0.689/0.718 & 0.467/0.465 & 0.526/0.540 & \textbf{0.710/0.725} & 0.398/0.384 & 0.630/0.656 \\
    KADID-10k~\cite{lin2019kadid} & 0.677/0.703 & 0.748/0.760 & 0.472/0.488 & 0.682/0.713 & 0.425/0.462 & \textbf{0.855/0.843} & 0.747/0.713 \\
    AGIQA-3k~\cite{li2023agiqa} &0.602/0.585 & 0.638/0.662 & 0.440/0.419 & 0.573/0.577 & 0.211/0.223 & 0.416/0.442 & \textbf{0.846/0.806} \\
    \bottomrule
    \end{tabular}%
    }
\label{tab:ablation_datasets}%
\end{table*}%

% \vspace{-3mm}
\noindent\textbf{Ablation study on the training strategy.} We conduct an ablation study on the proposed training strategy for \ours, with the results presented in Table \ref{tab:ablation_training}. In this table, \textit{MultiFunc$_{joint}$} corresponds to the straightforward way to build \ours~with a joint learning over different datasets. Nevertheless, it performs worse than \textit{MultiFunc$_{single}$} that is only trained on a single dataset, in terms of both within-dataset (on KonIQ-10k) or cross-dataset (on other datasets) quantitative performance. This observation experimentally reveals the inconformity issue in this field and indicates that ``the inconformity starves generalization'' as analyzed in Sec. \ref{sec:data_construction}. Moreover, we observe that the model \textit{Relat. + MultiFunc$_{joint}$} delivers clear improvements compared to the model \textit{MultiFunc$_{joint}$}, and the same goes for comparing \textit{Relat. + MultiFunc$_{single}$} to \textit{MultiFunc$_{single}$}. This experimental observation demonstrates the effectiveness of our applied relativity learning on utilizing diverse datasets with unaligned annotations in building an IQA generalist.
% an all-in-one foundation model. 
In specific, converting the absolute scores in each dataset into the judgements of relative quality can effectively make the model training free of the effects of the inconsistency in the original annotations of different datasets. Furthermore, the model \textit{Relat. + MultiFunc$_{single}$} performs the best across all model variants, demonstrating the superiority of our proposed training strategy for \ours. This experimental phenomenon provides an insight that the annotation inconsistency over different datasets inevitably affects the results of multifunctional training (Stage-2) as well. In the following, we study the effects of choosing different datasets when performing multifunctional training on the final results.

% \vspace{-3mm}
\noindent\textbf{Ablation study on the dataset choice.} Actually, in Table \ref{tab:quant_lmms}, only the performance on the KonIQ-10k dataset could not be affected much by the annotation inconformity issue across different datasets. For other datasets except KonIQ-10k, we only utilize the relativity of their data annotations for the Stage-1 training, without using their absolute numerical values. Therefore, the performance on these datasets could be considered to reflect a certain degree of generalization. Here, we zoom into the effects of the dataset choice used for the training of quantitative measurement in the Stage-2, denoted by the ``Source'' in the Table \ref{tab:ablation_datasets}. As the results show in Table \ref{tab:ablation_datasets}, for each test dataset denoted by ``Target'', the highest performance is reached when its corresponding training set is used for the training of quantitative measurement in the Stage-2. We find the data configuration in Table \ref{tab:dataset_and_training} achieves the best trade-off between different datasets when tackling the aforementioned annotation inconformity issue. Additionally, partially combining some of them for a joint training, instead of using them all, for \textit{MultiFunc$_{joint}$} in Table \ref{tab:ablation_training} is not the research focus of this work, we provide some results in the supplementary.

\vspace{-0.9em}
\begin{table}[!h]
\caption{Human side-by-side comparison of ours with other LLMs in qualitative Results.}
\centering
% \vspace{-0.6em}
\resizebox{0.8\linewidth}{!}{
\begin{tabular}{l|cccc}
\toprule
    \multicolumn{1}{c|}{G+S/S+B} & GPT-4V~\cite{gpt4vision} & LLaVA-v1.5~\cite{liu2023visual} & MiniGPT-4-13B~\cite{zhu2023minigpt} & InstructBLIP~\cite{instructblip}  \\ 
    \midrule
    \rowcolor[gray]{0.9}VisualCritic (Ours) v.s. & 1.29 & 1.77 & 1.82 & 2.45 \\
    \bottomrule
\hline
\end{tabular}
}
\vspace{-1.2em}
\label{tab:side-by-side}
\end{table}
% \vspace{-1.2em}

\subsection{Qualitative Results}
\label{sec:exp_qualitative}

We show the qualitative results in Figure \ref{fig:abstract_results}. From this case study, we can find that our proposed \ours~exhibits impressive instruction following capability with more detailed and pertinent descriptions compared to other LMMs, when being asked to assess the visual quality of the given image. In addition, other LMMs all misjudge the given AI-generated image as a photographic one while our \ours~makes the correct judgment. Besides these performances, \ours~also minimizes the generation of hallucinations and actively admits its unknown or uncertain judgments. More related qualitative results and their corresponding analysis are placed in our supplementary.

Quantitative evaluation of the linguistic outputs of LMMs remains an under-resolved issue so far. The common compromise solution is to employ a more advanced LMM (\egno, GPT-4V) as a critic for evaluating other models. But this is not adaptable to this work since our proposed model is at the forefront in this field. To address this issue as possible, we conduct a human side-by-side evaluation where 10 users are employed for comparing ours with 4 other LMMs on 100 random samples. The G+S/S+B scores (where \textbf{G}ood: our VisualCritic preferred, \textbf{S}ame: no preference, \textbf{B}ad: other model preferred) are as shown in Table~\ref{tab:side-by-side}.

\subsection{Prompts for Data Construction}
\label{sec: detail_prompt}

% \noindent\textbf{Unified format and data summary.} The training data for above three functionalities adopt a unified format following the conversation template design of Vicuna \cite{chiang2023vicuna} as below:
% \begin{tcolorbox}
% \textbf{Human:} $<img1>\{img1\_content\}</img1>(<\!img2\!>\{img2\_content\}<\!/img2\!>)[Task\;\\Identifier]\{instruction\_content\}$\\
% \textbf{Assistant:} $\{answer\_content\}$
% \end{tcolorbox}
% \noindent~In this format, $<img>$ and $</img>$ are special tokens marking the start and end of the token sequence of an image, respectively. $\{img\_content\}$, $\{instruction\_content\}$ and $\{answer\_content\}$ are three placeholders of the tokens of input image, user instruction and \ours's response, respectively. The contents in the brackets are optional, only existing in the training data of the relativity learning, not in others. Inspired by \cite{chen2023minigpt}, we treat the learning for aforementioned three functionalities as a multi-task problem and introduce a task identifier $[Task\;Identifier]$ to reduce the ambiguity across different tasks. It is instantiated as $<IQA\_QUANT>$, $<IQA\_DES>$ and $<AUTHENTICITY>$ for quantitative measurement, qualitative evaluation and authenticity detection, respectively. We summarize all adopted datasets for different purposes in Table \ref{tab:dataset_and_training}, and detail their 

\begin{figure*}[t]
	\centering
	\includegraphics[width=\textwidth]{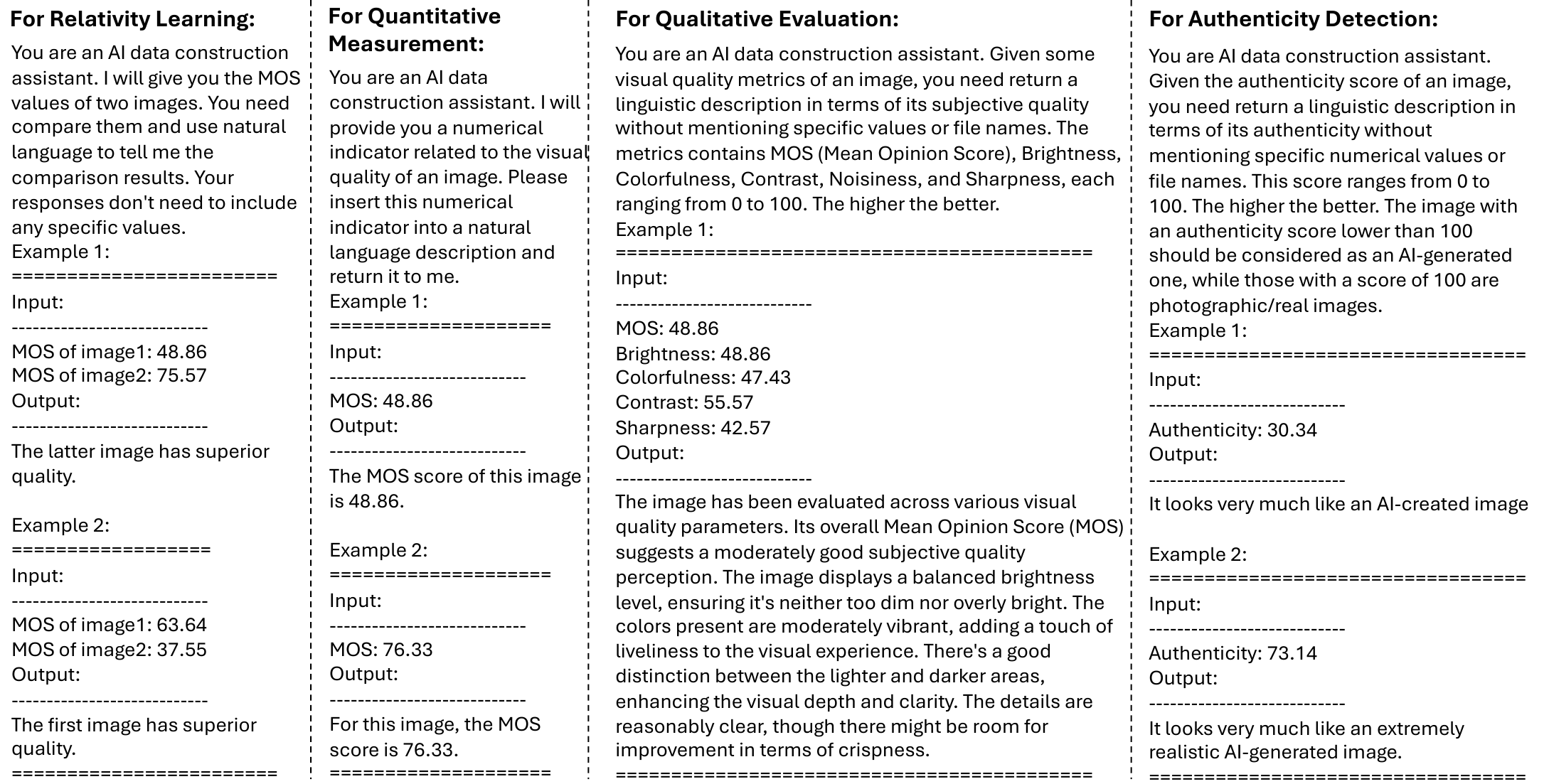}
        % \vspace{-6mm}
    	\caption{LLM prompts for data construction. These prompts from left to right are for generating \ours's response (\ieno, $\{answer\_content\}$) in the training data of relativity learning, quantitative measurement, qualitative evaluation and authenticity detection, respectively.}
	\label{fig:supp_prompts_answer}
\end{figure*}

As introduced in Sec.~\ref{sec:data_construction}, we prompt GPT-4 to automatically construct training data from public datasets with human subjective ratings, avoiding substantial manual annotation costs. The training data adopt a unified format as introduced in Sec.~\ref{box:unified_format}, where $\{img\_content\}$, $\{instruction\_content\}$ and $\{answer\_content\}$ are three placeholders of the tokens of input image, user instruction and \ours's response, respectively. The prompts to generate \ours's response (\ieno, $\{answer\_content\}$) are illustrated in Figure~\ref{fig:supp_prompts_answer} which are categorized into four segments, each corresponding to different image assessment functionalities task we investigate in Sec.~\ref{sec:data_construction}. In the supplementary, we provide additional prompts tailored for generating user instruction (\ieno, $\{instruction\_content\}$).

%%%%%%%%% CONCLUSION

\section{Conclusion}

In this work, we explore the question: \textit{Can LMMs perceive low-level visual quality aligned with human perception and show their excellent generalization ability in this regard?} We provide an answer with \ours, the first LMM of its kind for broad-spectrum image subjective quality assessment. Compared to conventional specialist models in the field of visual quality assessment, \ours~exhibits unprecedented cross-dataset generalizability and versatility. Beyond numerical scores, \ours~can further provide qualitative descriptions about visual quality and perform authenticity detection. We pinpoint the key to \ours~and introduce the core strategies on data construction and model training for making it real. We leave its extension to more data like videos and more functionalities in the future exploration. We also look forward to exploring its interplay with AIGC models for achieving better visual alignment.

\bibliographystyle{splncs04}
\bibliography{main}

\clearpage
\noindent\textbf{[Supplementary Material]}

In this supplementary material, we detail the experiment setup in Sec.~\ref{sec: exp_setup}, including the detailed introduction of training datasets in Sec.~\ref{sec: detail_dataset}, hyper-parameter configurations in Sec.~\ref{sec: detail_hyperparameter}, and more details about the adopted prompts for data construction in Sec.~\ref{sec: detail_prompt_supp}. We present more experiment results in Sec.~\ref{sec: exp_results}, including more quantitative results in Sec.~\ref{sec: result_within_dataset} and more qualitative results in Sec.~\ref{sec: result_qualitative}. Additionally, we present more discussion about the potential negative impact, limitations, and code release in Sec.~\ref{sec: more_discussion}.

\section{More Details about Experiment Setup}
\label{sec: exp_setup}

\subsection{Detailed Introduction of the Datasets}
\label{sec: detail_dataset}

In our main paper, we list all used datasets in Table 1. Aside from LLaVA-Instruction \cite{liu2023visual} and CCSBU-Aligned \cite{zhu2023minigpt} which are used for instruction tuning, the data for visual quality learning is multi-sourced. Among these datasets, KonIQ-10k \cite{hosu2020koniq}, SPAQ \cite{fang2020perceptual}, FLIVE \cite{ying2020patches}, LIVEC \cite{ghadiyaram2015massive} comprise photographic images with in-the-wild distortions. KADID-10k \cite{lin2019kadid} provide photographic images with artificial distortions. CGIQA-6K \cite{zhang2023subjective}, AGIQA-3k \cite{li2023agiqa} and AIGCIQA-2023 \cite{wang2023aigciqa2023} are all AI-generated images. More details about each one can be found in their corresponding publications. Owing to the substantial diversity over these datasets, the annotation inconformity introduced in our main paper is extremely pronounced, imposing a big challenge in the joint training across them. For all datasets, except the artificially distorted KADID-10k \cite{lin2019kadid}, we randomly sample 80\% of images as the training split while the remaining 20\% as the test split. The KADID-10k contains about 10k images, derived from 81 pristine images. Each image is degraded by 25 distortions in 5 levels. To ensure the diversity between training and test sets, we adopt all the distorted images corresponding to 65 of the pristine images as the training set, while the distorted images corresponding to the remaining 16 pristine images are used as the test set. Each training pair in Stage-1 is randomly sampled from the same dataset. 
Its detailed formulation is in Sec. 3.2 of the main paper and the specific prompts for its construction are in Figure 3 of the main paper.

\subsection{Hyper-parameter Configurations}
\label{sec: detail_hyperparameter}
We train~\ours~with Zero-2 powered by deepspeed framework~\cite{rajbhandari2021zero,rasley2020deepspeed}. The vision backbone has 1B parameters, and the vicuna model we used has 13B parameters, wherein the trainable parameters is $~$1B. The model is trained using a resolution of $224\times224$ at all stages. We employ the AdamW~\cite{loshchilov2017decoupled} optimizer paired with a cosine learning rate scheduler~\cite{loshchilov2016sgdr} for model training. In the first stage, we conduct the training for 100,000 steps on 4$\times$A100 GPUs, with a global batch size of 96 and a base learning rate of 2e-3. This stage is completed in approximately 30 hours. In the second stage, the model is trained for 20,000 steps on 4$\times$A100 GPUs, maintaining a global batch size of 96 and a base learning rate of 2e-4, taking around 6 hours. Finally, the model undergoes an additional 5,000 training steps on 4$\times$A100 GPUs, with a global batch size of 64, completed in about 1.5 hours. The maximum learning rate remains constant at 2e-5 during this final stage. We provide more detailed hyperparameter configurations in Table~\ref{tab:train_config}.

\begin{table}[htbp]
  \centering
    \caption{Details of the hyperparameter configurations.}
    \resizebox{0.5\textwidth}{!}{
    \begin{tabular}{l|ccc}
    \toprule
    \multicolumn{1}{c|}{Training Configuration} & Stage1 & Stage2 & Stage3 \\
    \midrule
    Base LR & 0.001 & 0.0001 & 0.00008 \\
    Min. LR & 0.0001 & 0.00001 & 0.000008 \\
    Warmup LR & 0.00001 & 0.00001 & 0.000001 \\
    Warm-up scheduler & Linear & Linear & Linear \\
    Scheduler & Cosine & Cosine & Cosine \\
    Weight decay & 0.01  & 0.05  & 0.05 \\
    Training iterations & 50000 & 150000 & 10000 \\
    Warmup iterations & 5000  & 10000 & 1000 \\
    LoRA\_r & 16    & 16    & 16 \\
    LoRA\_alpha & 32    & 32    & 32 \\
    LoRA\_dropout & 0.1   & 0.1   & 0.1 \\
    Batch size & 128   & 64    & 64 \\
    Iteration & 100000 & 50000 & 8000 \\
    Training Time & 20 h  & 10 h  & 1.6 h \\
    \bottomrule
    \end{tabular}%
    }
    % for each training stage.
    % \vspace{-3mm}
  \label{tab:train_config}%
\end{table}%

% During training, $\{instruction\_content\}$ in this format is randomly chosen from a predefined instruction set. Despite the limited number of instructions in this set, we experimentally find our proposed \ours~performs well with unseen instructions during inference. Such robust performance is achieved thanks to the generalizability of pre-trained LLMs and the role of instruction tuning. $\{answer\_content\}$ are generated by GPT-4 \cite{gpt4} via prompt engineering. We introduce the adopted core prompts as follows.
% For training, user instructions $\{instruction\_content\}$ and ground-truth responses $\{answer\_content\}$ are generated by GPT-4. Here, we illustrate the corresponding prompts for generating $\{answer\_content\}$ in the training data for relativity learning, quantitative measurement, qualitative evaluation and authenticity detection, respectively, in Figure \ref{fig:supp_prompts}.

\subsection{Prompts for Data Construction}
\label{sec: detail_prompt_supp}

As introduced in our main paper, the training data for enabling different assessment functionalities share a unified data format as below, where $\{img\_content\}$, $\{instruction\_content\}$ and $\{answer\_content\}$ are three placeholders of the tokens of input image, user instruction and \ours's response, respectively. 

\begin{tcolorbox}
\textbf{Human:} $<img1>\{img1\_content\}</img1>(<\!img2\!>\{img2\_content\}<\!/img2\!>)[Task\;\\Identifier]\{instruction\_content\}$\\
\textbf{Assistant:} $\{answer\_content\}$
\end{tcolorbox}

\begin{figure*}[t!]
% \begin{figure}[t]
    \vspace{-8mm}
	\centering
        % \flushleft
	\includegraphics[width=0.6\textwidth]{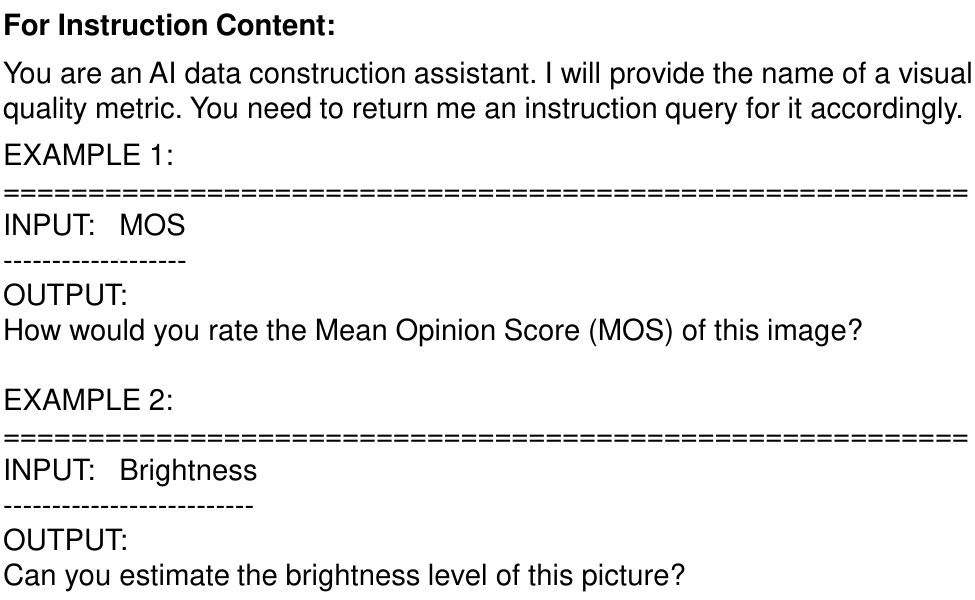}
        \caption{LLM prompts for data construction, \ieno, generating the $\{instruction\_content\}$ in the training data.}
    \vspace{-6mm}
	\label{fig:supp_prompts_instruct}
% \end{wrapfigure}
\end{figure*}

The prompts to generate $\{answer\_content\}$ are illustrated in Figure 3 in the main paper. The prompt generating user instruction (\ieno, $\{instruction\_content\}$) are shown in Figure~\ref{fig:supp_prompts_instruct}.

\section{More Experiment Results}
\label{sec: exp_results}

% \subsection{Comparison to conventional IQA models on within-dataset evaluation}


\subsection{More Quantitative Results}
\label{sec: result_within_dataset}

\noindent \textbf{Within-dataset comparison with IQA specialist models.} We have demonstrated the unprecedented generalizability of our proposed \ours~in Table 3 of our main paper. Here, we further compare \ours~to the state-of-the-art IQA specialist models in terms of their within-dataset quantitative performance. As shown in Table \ref{tab:quant_within_dataset}, our proposed \ours~delivers comparable within-dataset performance with those conventional IQA models, and even outperforms some of them. This is actually not easy since our \ours~is trained on diverse data with the aforementioned annotation inconformity issue over both photographic and AI-generated images. As a common sense in the IQA field, these conventional models, trained on an individual dataset, can easily fit one particular dataset but struggle to generalize to other datasets right out of box. Note that the generalizability and versatility are the primary exploration objectives of this paper, while the within-domain performance is not. We thus place these results in the supplementary material and leave the improvements in the balance of within-dataset and cross-dataset performance in our future studies.

\begin{table*}[th!]
  \centering
  \caption{Comparison results (MOS) of our \ours~with the state-of-the-art IQA specialist models for within-dataset settings.}
    \resizebox{0.75\textwidth}{!}{
    \begin{tabular}{l|ccccc}
    \toprule
    \multicolumn{1}{c|}{Dataset}      & KonIQ-10k & SPAQ  & LIVEC & KADID10k & FLIVE \\
    \midrule
    BRISQUE~\cite{mittal2012no} & 0.665/0.681 & 0.809/0.817 & 0.608/0.629 & 0.528/0.567 & 0.288/0.373 \\
    CORNIA~\cite{ye2012unsupervised} & 0.780/0.795 & 0.709/0.725 & 0.629/0.671 &  0.516/0.558 & -  \\
    ILNIQE~\cite{zhang2015feature}  & 0.507/0.523 &   -   & 0.432/0.508 &   -    & -  \\
    DBCNN~\cite{zhang2018blind} & 0.875/0.884  &  0.911/0.915 & 0.851/0.869 & 0.851/0.856 & 0.580/0.641 \\
    CONTRIQUE~\cite{madhusudana2022image}  & 0.894/0.906 & 0.914/0.919 &  0.845/0.857 & 0.934/0.937 & 0.580/0.641 \\
    MUSIQ~\cite{ke2021musiq} & 0.916/0.928 & 0.917/0.921 &   -    &    -   & 0.646/0.739 \\
    Re-IQA~\cite{saha2023re} & 0.914 0.923 & 0.918/0.925 & 0.840/0.854 & 0.872/0.885 & 0.645/0.733 \\
    \midrule
    % Ours(single) & 0.787/0.803 & 0.805/0.791 & 0.804/0.834 & 0.821/0.838 & - \\
    \rowcolor[gray]{0.9}\ours~ & 0.856/0.867 & 0.865/0.864 & 0.857/0.864 & 0.855/0.843 & 0.633/0.667 \\
    \bottomrule
    \end{tabular}%
    }
  \label{tab:quant_within_dataset}%
\end{table*}%

\noindent \textbf{The accuracy of relativity learning.} As introduced in Sec 3.2 and Table 5 in the main paper, relativity learning is an important part of our \ours, we show the accuracy of this in Table~\ref{tab:relative_learning}.  As the results shown, judging a pair with ranking is slightly better than directly doing A-B on their scores. 

% \vspace{-0.79em}
\begin{table}[!h]
\centering
\caption{Comparison the accuracy of relativity learning with directly doing A-B on their output scores.}
% \vspace{-0.6em}
\resizebox{0.65\textwidth}{!}{
\begin{tabular}{l|ccccccc}
\toprule
Methods & KonIQ & SPAQ & FLIVE & LIVEC & CGIQA & KADID & AGIQA \\
\midrule
Ranking & 0.85  & 0.86  & 0.76  & 0.80  & 0.74  & 0.79 & 0.82 \\
A-B scores & 0.81  & 0.84  & 0.72  & 0.76  & 0.70   & 0.77 & 0.80 \\
\bottomrule
\end{tabular}
}
% \vspace{-1.2em}
% \caption{Relative pairs test acccuracy.}
\label{tab:relative_learning}
\end{table}
% \vspace{-1.2em}

\noindent \textbf{Combining 2 or 3 training datasets.} 
As illustrated in Table 3 of the main paper, \textit{MultiFunc$_{single}$} significantly outperforms \textit{MultiFunc$_{joint}$}. This demonstrates the existence of the aforementioned annotation inconformity issue and presents a solution. While there are alternative approaches, such as combining two or three datasets instead of all seven, these methods do not fundamentally differ from using all seven datasets (\textit{MultiFunc$_{joint}$}) because they still encounter the annotation inconformity issue. Furthermore, exhaustively training on all $127$ combinations from the sum $\sum_{k=1}^{7} C(7, k)$ to seek the highest performing structure is not only impractical due to the computational resources required, but it also falls outside the scope of this paper.
We have provided a comparison with combinations of two or three datasets in Table~\ref{tab:multiple_joint} for reference. Our experimental results have not shown any performance improvements over experiments conducted with single training datasets.

\begin{table}[!h]
\small
\centering
% \vspace{-0.6em}
\caption{Comparison of models trained with combinations of two and three datasets versus models trained with single datasets in the Stage-2. `Two' denotes `KonIQ+FLIVE', `Three' represents `KonIQ+SPAQ+CGIQA' and `Single' means `KonIQ'.}
\resizebox{\textwidth}{!}{
\begin{tabular}{l|ccccccc}
 \toprule
 Datasets. & KonIQ & SPAQ & FLIVE & LIVEC & CGIQA & KADID & AGIQA \\ 
 \midrule
    Two & 0.722/0.741 & 0.766/0.789 & 0.467/0.485 & 0.638/0.664 & 0.334/0.357 & 0.507/0.521 & 0.644/0.668 \\
    Three  & 0.694/0.733 & 0.723/0.736 & 0.288/0.314 & 0.553/0.579 & 0.294/0.301 & 0.443/0.496 & 0.584/0.563 \\
    \rowcolor[gray]{0.9}Single & \textbf{0.856/0.867} & \textbf{0.844/0.852} & \textbf{0.553/0.633} & \textbf{0.738/0.778} & \textbf{0.442/0.473} & \textbf{0.681/0.701} & \textbf{0.734/0.712} \\
 \bottomrule
\end{tabular}
}
% \vspace{-1.2em}
\label{tab:multiple_joint}
\end{table}
\vspace{-1.2em}

\subsection{More Qualitative Results}
\label{sec: result_qualitative}

\begin{figure*}[t!]
	\centering
	\includegraphics[width=\textwidth]{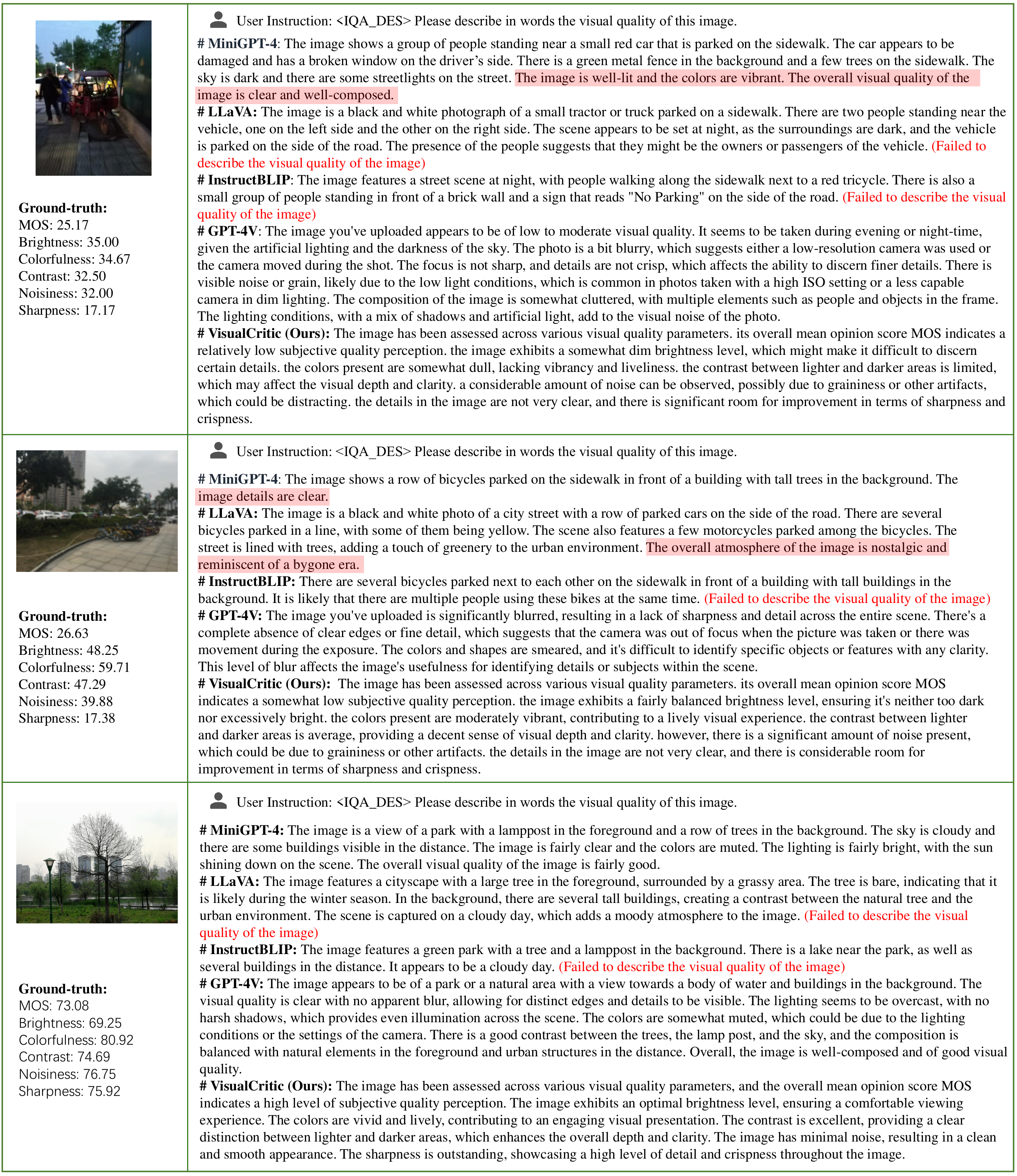}
    \caption{More comparison results between our \ours~and other LMMs on qualitative evaluation. Errors are red-highlighted.}
	\label{fig:supp_qualitative}
\end{figure*}

\begin{figure*}[t!]
	\centering
	\includegraphics[width=\textwidth]{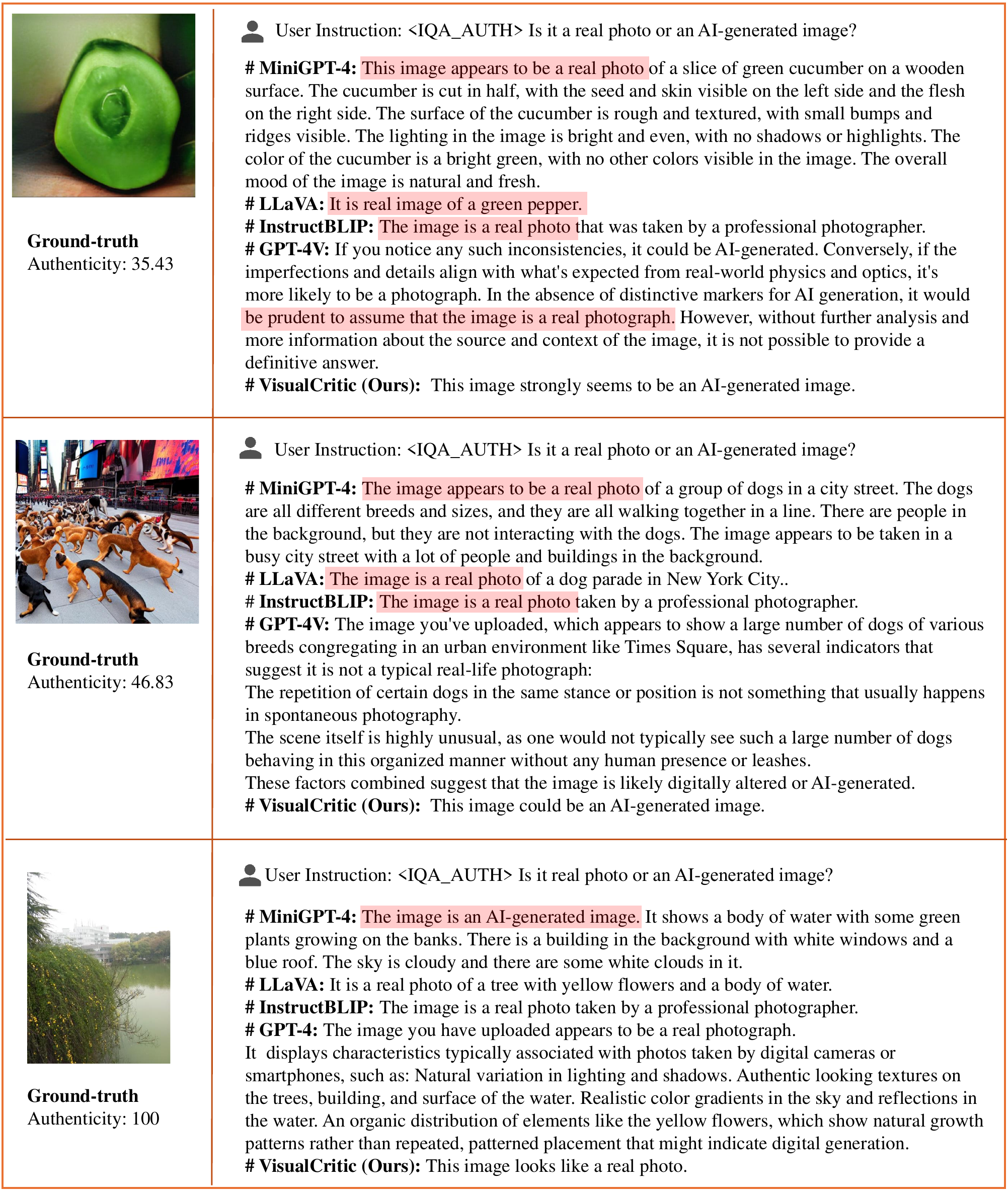}
    \caption{More comparison results between our \ours~and other LMMs on authenticity detection. Errors are red-highlighted.}
	\label{fig:supp_authenticity}
\end{figure*}

\begin{figure*}[t!]
	\centering
	\includegraphics[width=\textwidth]{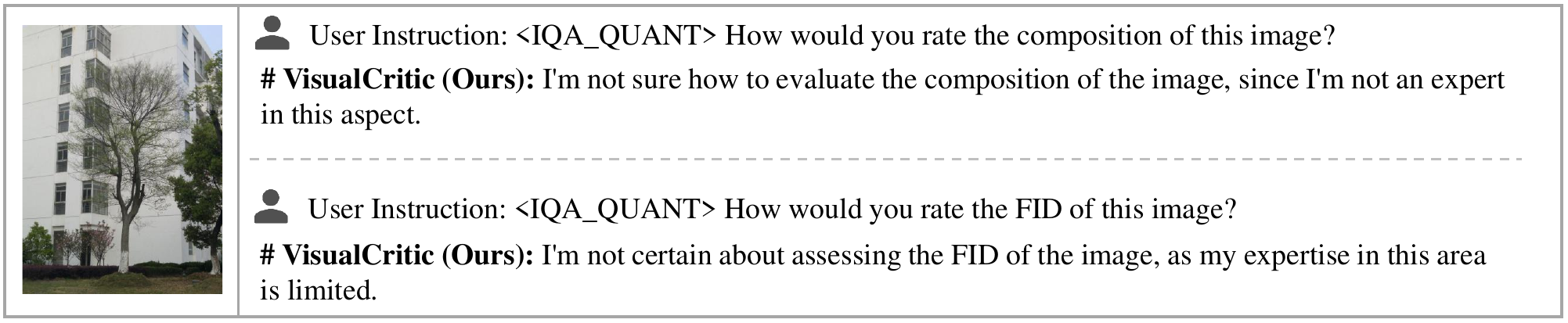}
    \caption{Illustration of the answers given by our proposed \ours~with unknown or uncertain judgements.}
	\label{fig:supp_negative}
\end{figure*}

\noindent \textbf{More results on qualitative evaluation}
We have illustrated the comparison results between \ours~and other LMMs regarding qualitatively commenting on the visual quality of a given image in our main paper. Here, we present more in Figure \ref{fig:supp_qualitative}. In this regard, MiniGPT-4 \cite{zhu2023minigpt} seems to have a bias of considering all images as the ones with high visual quality. LLaVA \cite{liu2023visual} and InstructBLIP \cite{instructblip} often provide responses that are irrelevant to the instructions, that is, their responses do not contain any content related to visual quality. GPT-4V \cite{yang2023dawn} can give basically reasonable answers in this regard but performs poorly in quantitative measurement and authenticity detection. They all include the contents related to semantic understanding in their responses, while our \ours~is more focused on the visual quality in line with the instructions.

\noindent \textbf{More results on authenticity detection}
We present more results on the case studies regarding authenticity detection in Figure \ref{fig:supp_authenticity}. As shown in this figure, the open-source LMMs MiniGPT-4 \cite{zhu2023minigpt}, LLaVA \cite{liu2023visual} and InstructBLIP \cite{instructblip} all tend to regard the given images as real/photographic, whether these images are actually real or AI-generated. Our proposed \ours~provides the most accurate judgements, aligned with the ground-truths. Besides, the degree of certainty for these judgements can also be reflected in their responses.

\noindent \textbf{Answers with unknown or uncertain judgements}
Besides our targeted generalizability and versatility, we also make additional efforts in minimizing the generation of hallucinations and actively admits its unknown or uncertain judgments. To achieve this, we add extra carefully designed training samples to enable \ours~to inform users when user instructions are out of the range that its knowledge currently covers. We showcase the performance of \ours~in this kind of scenarios in Figure \ref{fig:supp_negative}. Please note that this is an additional point we consider for building a practical system, not as the focus of this research paper.

\section{More discussion}
\label{sec: more_discussion}

\noindent \textbf{Potential negative impact.} Job displacement may arise as automated assessments reduce the demand for human expertise in visual assessment industries. Additionally, overreliance on the LVLM for image quality assessments may lead to a devaluation of human judgment, risking the neglect of nuanced analysis that AI cannot replicate.

\noindent \textbf{Limitation.} The limitation mainly lies in that all training data of this work come from existing IQA datasets. Collecting more high-quality labeled data and extending this work to other types of data (\egno, audio, 3D data, \etc), while beyond the current research scope, is very worthwhile to pursue in the future. 

\noindent \textbf{Code release.} Our data, code, and models will be released upon paper acceptance.

\end{document}